\documentclass[a4paper]{article}

\usepackage{INTERSPEECH2022}
\usepackage{url}

\DeclareMathOperator*{\argmax}{arg\,max}

\title{Non-autoregressive Error Correction for CTC-based ASR \\ with Phone-conditioned Masked LM}

\name{Hayato Futami$^1$, Hirofumi Inaguma$^1$, Sei Ueno$^1$, Masato Mimura$^1$, \\ Shinsuke Sakai$^1$, Tatsuya Kawahara$^1$}
\address{
  $^1$Graduate School of Informatics, Kyoto University, Sakyo-ku, Kyoto, Japan}
\email{\{surname\}@sap.ist.i.kyoto-u.ac.jp}

\begin{document}

\maketitle
\begin{abstract}
Connectionist temporal classification (CTC) -based models are attractive in automatic speech recognition (ASR) because of their non-autoregressive nature.
To take advantage of text-only data, language model (LM) integration approaches such as rescoring and shallow fusion have been widely used for CTC.
However, they lose CTC's non-autoregressive nature because of the need for beam search, which slows down the inference speed.
In this study, we propose an error correction method with phone-conditioned masked LM (PC-MLM).
In the proposed method, less confident word tokens in a greedy decoded output from CTC are masked.
PC-MLM then predicts these masked word tokens given unmasked words and phones supplementally predicted from CTC.
We further extend it to Deletable PC-MLM in order to address insertion errors.
Since both CTC and PC-MLM are non-autoregressive models, the method enables fast LM integration.
Experimental evaluations on the Corpus of Spontaneous Japanese (CSJ) and TED-LIUM2 in domain adaptation setting shows that our proposed method outperformed rescoring and shallow fusion in terms of inference speed, and also in terms of recognition accuracy on CSJ.
\end{abstract}

\noindent\textbf{Index Terms}: speech recognition, CTC, language model, error correction

\vspace{-2pt}
\section{Introduction}
\vspace{-2pt}
End-to-end automatic speech recognition (ASR) that directly maps acoustic features into text sequences has shown remarkable results.
For its modeling, CTC-based models \cite{Graves06-CTC}, attention-based sequence-to-sequence models \cite{Chan16-LAS, Dong18-ST}, and neural network transducers \cite{Graves12-ST, Zhang20-TT} are mainly used.
Among them, CTC-based models have the advantage of lightweight and fast inference, which we focus on in this study.
They consist of an encoder followed by a compact linear layer only and can predict all tokens in parallel, which is called non-autoregressive generation.
Attention-based models and transducers predict tokens one-by-one depending on previously decoded tokens, which is called autoregressive generation.

End-to-end ASR models including CTC-based models are trained on paired speech and transcripts.
However, it is usually difficult to prepare a sufficient amount of paired data on target domain.
On the other hand, much larger amount of in-domain text-only data is often available, and the most popular way to leverage it in end-to-end ASR is the integration of external language models (LMs).
In rescoring \cite{Mikolov10-RNN, Shin19-ESS, Salazar20-MLMS}, $n$-best hypotheses obtained from an ASR model are rescored by an LM, and then the hypothesis of the highest score is selected.
In shallow fusion \cite{Chorowski17-TBD, Kannan18-AILM}, the interpolated score of the ASR model and the LM is calculated at each ASR decoding step.
These two LM integration approaches are simple and effective, and therefore they are widely used in CTC-based ASR.
However, they deteriorate the fast inference, which is the most important advantage of CTC over other variants of end-to-end ASR.
Specifically, beam search to obtain multiple hypotheses makes CTC lose its non-autoregressive nature \cite{Graves14-TES}.

Besides rescoring and shallow fusion, knowledge distillation (KD) \cite{Hinton15-KD} -based LM integration for attention-based ASR has been proposed \cite{Bai19-LST, Futami20-DKB, Bai21-FESR}.
The knowledge of the LM (teacher model) is transferred to the ASR model (student model) during ASR training.
Though KD has the advantage of no additional inference steps during testing, the effect of LM is often limited because LM cannot directly affect the ASR inference.

In this study, we propose an ASR error correction method where a masked LM (MLM) corrects less confident tokens in a CTC-based ASR hypothesis.
This method does not require beam search in CTC and corrects all less confident parts in parallel using MLM.
In other words, both ASR and error correction procedures are conducted in a non-autoregressive manner.
However, this MLM-based error correction does not work well because MLM does not consider acoustic information in correction.
To solve the problem, we propose phone-conditioned masked LM (PC-MLM) to leverage phone information.
CTC is jointly trained to predict phones from the intermediate layer of its encoder noted as hierarchical multi-task learning \cite{Krishna18-HMTL}.
PC-MLM uses both word and phone context for correction.
Furthermore, to deal with insertion errors, we propose Deletable PC-MLM that is trained to predict insertions to be removed.

\vspace{-2pt}
\section{Preliminaries and related work}
\vspace{-2pt}

\subsection{CTC-based ASR}
Let $\bm{X} = (\bm{x}_1, ..., \bm{x}_t, ..., \bm{x}_T)$ denote the acoustic features and $\bm{y} = (y_1, ..., y_i, ..., y_L)$ denote the label sequence of tokens corresponding to $\bm{X}$.
An encoder network transforms $\bm{X}$ into a higher-level representation of length $T'$.
A CTC-based model predicts frame-level CTC path $\bm{\pi} = (\pi_1, ..., \pi_{T'})$ using the encoded representations.
Let $\mathcal{V}$ denote the vocabulary and $\phi$ denote a blank token.
We define the probability of predicting $v \in \mathcal{V} \cup \{\phi\}$ for the $t$-th time frame as
\begin{align}
\label{eq:prob-ctc}
P_{\rm CTC}^{(t, v)} = p(v \, | \, \bm{X}, t).
\end{align}
In greedy decoding, CTC path $\bm{\pi}$ is decided as
\begin{align}
\label{eq:path-ctc}
\pi_t = \argmax_v P_{\rm CTC}^{(t, v)},
\end{align}
which is based on non-autoregressive generation.
In beam search decoding \cite{Graves14-TES}, $\pi_t$ is dependent on previously decided path $\pi_{<t} = (\pi_1, ..., \pi_{t-1})$ and its score, which is autoregressive.
The output sequence $\bm{y}$ is obtained by $\bm{y} = \mathcal{B}(\bm{\pi})$, where the mapping $\mathcal{B}$ removes blank tokens after condensing repeated tokens.

The CTC loss function is defined over all possible paths that can be reduced to $\bm{y}$:
\begin{align}
\label{eq:loss-ctc}
\mathcal{L}_{\rm CTC} = - \log p(\bm{y} | \bm{X}) = - \sum_{\bm{\pi} \in \mathcal{B}^{-1}(\bm{y})} \log p(\bm{\pi} | \bm{X}).
\end{align}

\subsection{Masked LM}
Masked LM (MLM) is originally proposed as a pre-training objective for BERT \cite{Devlin19-BERT}, which has shown promising results in many downstream NLP tasks.
During MLM training, some of the input tokens (usually $15\%$) are masked and the original tokens are predicted.
MLM predicts all masked tokens in parallel, or non-autoregressively, given both left and right unmasked context.
We define the probability of predicting $v \in \mathcal{V}$ for the $i$-th token as
\begin{align}
P_{\rm MLM}^{(i, v)} &= p(v\,|\, \bm{y}^{\rm (mask)}, i),
\end{align}
where the $i$-th token is masked in $\bm{y}^{\rm (mask)}$.
Conventionally, RNN or Transformer LMs has been used in ASR via rescoring \cite{Mikolov10-RNN} and shallow fusion \cite{Chorowski17-TBD, Kannan18-AILM}.
They predicts each token autoregressively, given only their left context.
Recently, MLM has been applied to ASR via rescoring \cite{Shin19-ESS, Salazar20-MLMS} and knowledge distillation (KD) \cite{Futami20-DKB, Bai21-FESR}.
MLM has been reported to perform better than conventional LMs thanks to the use of the bidirectional context.
However, rescoring with MLM takes a lot of time during testing because it requires $L$ steps to rescore a hypothesis of length $L$ by masking each token \cite{Salazar20-MLMS, Futami21-ELECTRA}.
In KD with MLM, the following KL-divergence based objective is minimized during ASR training.
\begin{align}
\label{eq:L_KD-att}
\mathcal{L}_{\rm KD} = - \sum_{i=1}^{L} \sum_{v \in \mathcal{V}} P_{\rm MLM}^{(i, v)} \log P_{\rm Att}^{(i, v)},
\end{align}
where $P_{\rm Att}^{(i, v)}$ denotes the probability of predicting $v$ for the $i$-th token with an attention-based ASR model.
In the previous studies, the student ASR model has been limited to an attention-based model that makes token-level predictions as MLM.

As an extension of MLM, conditional masked LM (CMLM) has been proposed in \cite{Ghazvininejad19-MP} for non-autoregressive neural machine translation (NMT).
CMLM is an encoder-decoder model that predicts all masked tokens in a non-autoregressive manner, conditioning on both source text and unmasked target translation.
In ASR, Audio-CMLM (A-CMLM) \cite{Chen19-LAF} and mask CTC \cite{Higuchi20-MCTC} adopts CMLM architecture for non-autoregressive ASR.
In mask CTC, similar to our proposed method, less confident tokens in CTC output are refined by CMLM.
However, it conditions on acoustic features and is jointly trained with CTC on paired data, while our proposed PC-MLM conditions on phone tokens and separately trained from CTC on text-only data.

\subsection{ASR error correction}
ASR error correction aims to correct errors generated by ASR using another high-level model.
Recently, it has been modeled with autoregressive sequence-to-sequence models that convert an ASR hypothesis to a corrected one, like neural machine translation \cite{Guo19-SC, Zhang19-ITSC, Mani20-AECDA, Wang20-AECAT, Hrinchuk20-CASR, Zhao21-BART}.
They are usually trained on paired ASR hypotheses and their corresponding references.
However, such paired data is obtained from a limited amount of paired speech and transcripts, which can cause an overfitting problem.
Some studies \cite{Hrinchuk20-CASR, Zhao21-BART} use text-only data via initialization with a large pre-trained LM such as BERT and BART \cite{Lewis19-BART}.
In \cite{Guo19-SC}, recognition results of TTS-synthesized speech are used as pseudo ASR hypotheses, and in \cite{Wang20-AECAT}, text-level simulated errors based on n-gram confusion matrix are used.
In \cite{Wang20-AECAT}, a phone-level encoder is added to a sequence-to-sequence model to incorporate phone information.
More recently, a non-autoregressive error correction model based on edit alignment was proposed in \cite{Leng21-FC}.
In this study, phone-conditioned masked LM is used as an error correction model.
It is trained not on paired data but on text-only data and realizes non-autoregressive and phone-aware correction.

\vspace{-2pt}
\section{Proposed method}
\vspace{-2pt}
\subsection{Phone-conditioned masked LM (PC-MLM)}
Phone-conditioned masked LM (PC-MLM) is a phone-to-word conversion model that consists of a Transformer-based CMLM \cite{Ghazvininejad19-MP}.
PC-MLM predicts word tokens of the masked positions given both phone tokens $\bm{p}$ input to the encoder and word tokens $\bm{y}^{\rm (mask)}$ input to the decoder.
When the $i$-th token is masked in $\bm{y}^{\rm (mask)}$, the probability of predicting $v \in \mathcal{V}$ for the $i$-th token can be defined as
\begin{align}
\label{eq:prob-pcmlm}
P_{\rm PC-MLM}^{(i, v)} &= p(v\,|\, \bm{p}, \bm{y}^{\rm (mask)}, i).
\end{align}
Phone information can be automatically obtained from word sequences using a lexicon, so PC-MLM can be trained on text-only data as LMs.
To prevent overfitting, some phone tokens are randomly masked ($20\%$) during training, which is called ``text augmentation'' in \cite{Wang21-CRNNT}.

\subsection{Error correction with PC-MLM}
\label{sec-proposed-ec}
\begin{figure}[t]
  \centering
  \fbox{
  \includegraphics[width=0.95\linewidth]{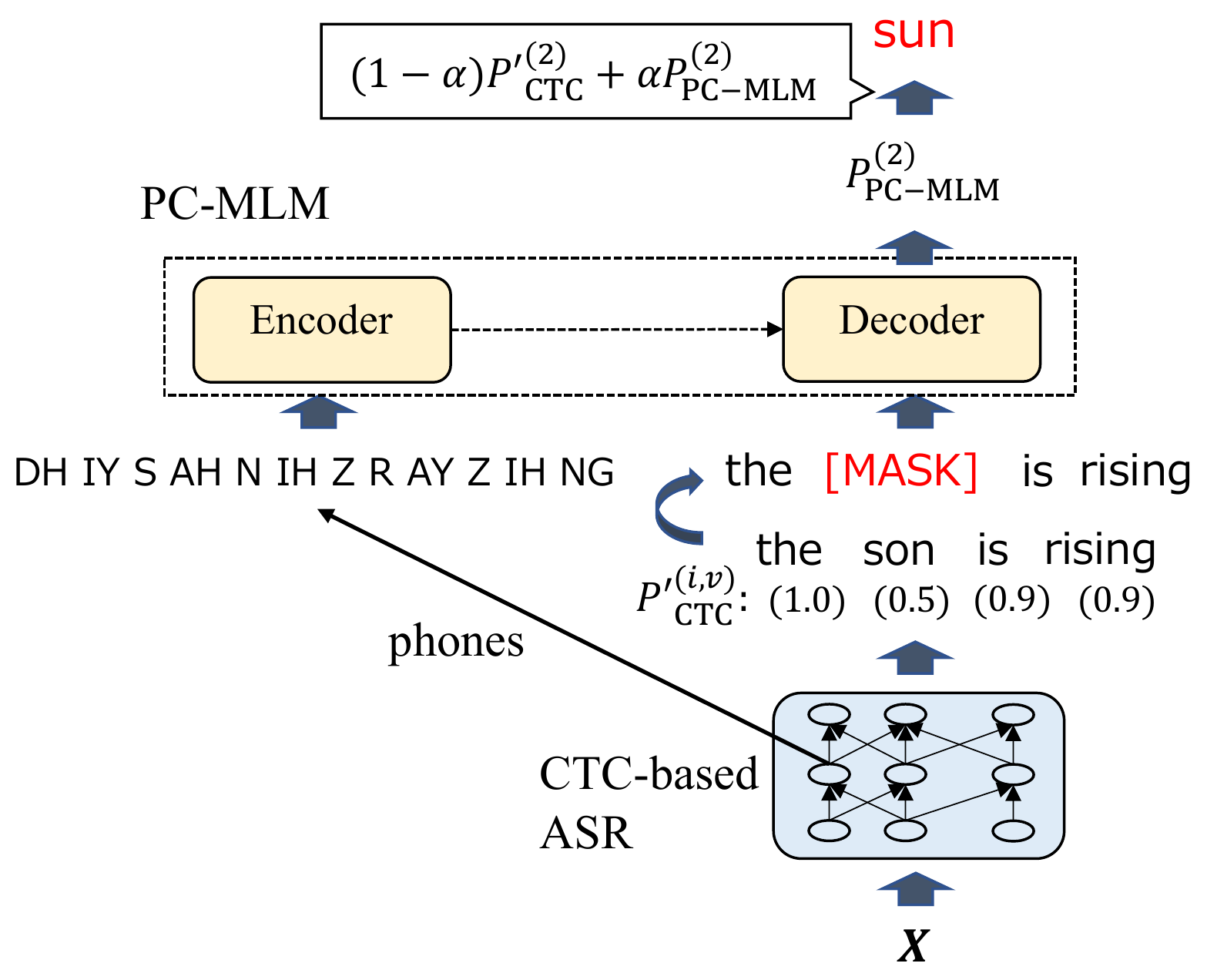}
  }
  \caption{Illustration of proposed error correction with PC-MLM for CTC-based ASR. The threshold $\beta$ in Eq. (\ref{eq:mask-threshold}) is set to $0.8$ in the figure, and therefore ``son'' of $P'^{(i, v)}_{\rm CTC} < 0.8$ is masked and then corrected by PC-MLM.}
  \label{fig:overview-ec}
\end{figure}

In this study, PC-MLM serves as an error correction model that corrects CTC-based ASR hypotheses.
An overview of the proposed method is illustrated in Figure \ref{fig:overview-ec}.
We use confidence scores to determine which tokens are to be masked and then corrected.
First, to obtain token-level confidence scores, we need to aggregate frame-level CTC predictions in Eq. (\ref{eq:prob-ctc}) into token-level predictions as
\begin{align}
\mathcal{A}(i) &= \argmax_t P^{(t, y_i)}_{\rm CTC} \, {\rm s.t.} \, \pi_t = y_i, \\
P'^{(i, \cdot)}_{\rm CTC} &= P^{(\mathcal{A}(i), \cdot)}_{\rm CTC},
\end{align}
where the index mapping $\mathcal{A}$ from $i$ to $t$ is obtained from greedy CTC path $\bm{\pi}$ in Eq. (\ref{eq:path-ctc}).
Then, a part of CTC output $\bm{y}$ is masked out to obtain $\bm{y}^{\rm (mask)}$ based on the confidence score as
\begin{align}
\label{eq:mask-threshold}
y^{\rm (mask)}_i =
\begin{cases}
\text{\url{[MASK]}} & P'^{(i, y_i)}_{\rm CTC} < \beta \\
y_i & P'^{(i, y_i)}_{\rm CTC} \geq \beta.
\end{cases}
\end{align}

In addition to word-level context $\bm{y}^{\rm (mask)}$, we propose to leverage phone-level context $\bm{p}$ in error correction as input to PC-MLM.
We obtain these phone-level predictions via a hierarchical multi-task learning framework \cite{Krishna18-HMTL}, where an auxiliary phone-level target is added at an intermediate layer of the encoder.
This method also improves word-level ASR at the final layer.

Then, as in Eq. (\ref{eq:prob-pcmlm}), PC-MLM provides $P_{\rm PC-MLM}^{(i, v)}$ given $\bm{y}^{\rm (mask)}$ and $\bm{p}$.
Finally, to get a corrected hypothesis $\bm{y}^{\rm (correct)}$, we can directly use the probability of PC-MLM for correction.
We also propose to use the interpolated score of CTC and PC-MLM as
\begin{align}
\label{eq:ec-score-inter}
y^{\rm (correct)}_i = \argmax_v ((1 - \alpha) P_{\rm CTC}^{'(i, v)} + \alpha P_{\rm PC-MLM}^{(i, v)}).
\end{align}

The proposed method has an advantage of fast inference over existing LM integration methods for CTC such as rescoring and shallow fusion.
Our method requires only a $1$-best hypothesis, while rescoring or shallow fusion requires $n$-best ($n > 1$) hypotheses after or during decoding.
For CTC, a $1$-best one can be obtained by non-autoregressive generation, but $n$-best ones can be obtained by autoregressive generation, whose difference has a significant impact on inference speed.
After obtaining hypotheses, PC-MLM corrects tokens in a non-autoregressive manner, which is fast.
Our method can also be applied to decoded outputs from attention models and transducers.
Furthermore, our method can apply LM to non-autoregressive models other than CTC such as A-CMLM \cite{Chen19-LAF}, LASO \cite{Bai20-LASO}, Insertion Transformer \cite{Fujita20-IBM} that had difficulty in LM integration because they do not have beam search.

\subsection{Deletable PC-MLM}
PC-MLM is trained to replace masked tokens with the same number of other tokens, which only deal with substitution errors from CTC-based ASR.
We further propose Deletable PC-MLM to address insertion errors, inspired by \cite{Gu19-LT, Higuchi21-IMCTC}.
Deletable PC-MLM predicts null tokens ($\phi$) for inserted errors, and then a corrected result $\bm{y}^{\rm (correct)}$ is obtained by removing them.
During its training, some of the input tokens are randomly masked ($15\%$) as in MLM, and some mask tokens (\url{[MASK]}) are randomly inserted between them.
The number of mask tokens to be inserted is sampled from Poisson distribution ($\lambda = 0.2$) as in \cite{Lewis19-BART}.
After masking and insertion, Deletable PC-MLM is trained to predict original tokens from non-inserted positions and null tokens from inserted positions.

\vspace{-2pt}
\section{Experimental evaluations}
\vspace{-2pt}
\subsection{Experimental conditions}
We evaluated our methods on ASR using the Corpus of Spontaneous Japanese (CSJ) \cite{maekawa03-CSJ} and the TED-LIUM2 corpus \cite{Ted214}.
CSJ consists of Japanese presentations, including CSJ-APS subcorpus ($240$ hours) on academic presentation speech and CSJ-SPS subcorpus ($280$ hours) on simulated public speaking on everyday topics.
TED-LIUM2 consists of English presentations available on the TED website.
Evaluations were done in a domain adaptation setting, where we assumed that paired data for training ASR on target domain was not available and that text-only data on target domain was available.
An ASR model was trained on paired data on another source domain and evaluated on target domain.
In CSJ experiments, the ASR model was trained on CSJ-SPS (source domain) and evaluated on the test set for CSJ-APS (target domain), and LMs were trained on CSJ-APS transcripts.
In TED-LIUM2 experiments, we adopted Librispeech \cite{Libri15} corpus for source domain, which has $960$ hours of paired data on English book reading.
An ASR model was trained on Librispeech (source domain) and evaluated on the test set for TED-LIUM2 (target domain), and LMs were trained on $250$M-word text prepared for TED-LIUM2 \cite{Ted214}.

We prepared a CTC-based ASR model that consists of a Transformer encoder with $L=12, H=256, A=4$ and a linear layer, where $L$, $H$ and $A$ denotes the number of layers, hidden units, and attention heads, respectively.
We also prepared a Conformer \cite{Gulati20-CF} -based one with the same size.
We used Adam optimizer with Noam learning rate scheduling \cite{Dong18-ST} of $warmup\_n = 25,000, k = 5$.
For data augmentation, SpecAugment \cite{Park19-SA} was applied to acoustic features.
We prepared four types of language models (LMs): Transformer LM (TLM), masked LM (MLM), phone-conditioned masked LM (PC-MLM), and Deletable PC-MLM (Del PC-MLM).
TLM and MLM share the same architecture that consists of a Transformer encoder with $L=12, H=256, A=4$.
PC-MLMs consist of a Transformer encoder and decoder with $L=4, H=256, A=4$, which has almost the same number of parameters as TLM and MLM.
We used Adam optimizer of the learning rate of $10^{-4}$ with learning rate warmup over the first $10\%$ of total steps and linear decay for TLM and MLM.
For PC-MLMs, Noam learning rate scheduling was applied.

ASR and LMs shared the same Byte Pair Encoding (BPE) subword vocabulary for each corpus.
The BPE vocabulary of $10872$ and $9798$ entries were used in CSJ and TED-LIUM2 experiments, respectively.
For PC-MLM, phone tokens were obtained using OpenJTalk-based grapheme-to-phone (g2p) tool \footnote{\url{https://github.com/r9y9/pyopenjtalk}} for CSJ ($45$ entries), and officially provided pronunciation dictionaries for TED-LIUM2 ($44$ entries).
All our implementations are publicly available \footnote{\url{https://github.com/emonosuke/emoASR/tree/main/asr/correct}}.

\subsection{Experimental results}
Table \ref{tab:ec-csj} shows the ASR results of our proposed error correction (EC) on CSJ.
As mentioned in Section \ref{sec-proposed-ec}, a baseline CTC-based model \url{(A1)} is trained with hierarchical multi-task learning \cite{Krishna18-HMTL}, which was confirmed to improve the word error rate (WER) from $18.44\%$ to $18.10\%$ with $9.10\%$ phone error rate (PER).
We compared three LMs for error correction: MLM, PC-MLM, and Del PC-MLM.
We also compared ``with'' and ``without'' score interpolation in Eq. (\ref{eq:ec-score-inter}), where $\alpha = 1$ in the table means ``without'' interpolation.
With score interpolation ($\alpha < 1$), $\alpha$ was determined using the development set, and its value is shown in the table. 
First, MLM that ignores phone information did not perform well, even with score interpolation \url{(A2, A3)}.
PC-MLM that considers phone information outperformed the baseline \url{(A4)}, and the score interpolation led to further improvement \url{(A5)}.
Del PC-MLM that is trained to delete insertion errors further improved WER \url{(A6, A7)}.
Compared to PC-MLM \url{(A5)}, Del PC-MLM \url{(A7)} actually reduced insertion errors from $623$ to $363$ together with substitution errors from $2976$ to $2660$, but increased deletion errors from $811$ to $1266$.
The cascade approach with phone-to-word CTC \url{(A8)} can be considered, but it did not perform well because of error propagation.
We saw that our method was also effective for an improved Conformer-based baseline \url{(B1, B2)}.

\begingroup
\renewcommand{\arraystretch}{1.1}
\begin{table}[t]
  \caption{ASR results on CSJ-APS with proposed error correction (EC). ASR models were trained on CSJ-SPS, and LMs were trained on CSJ-APS transcripts. $\alpha$ denotes the interpolation weight in Eq. (\ref{eq:ec-score-inter}), and $\alpha = 1$ means no score interpolation.}
  \label{tab:ec-csj}
  \centering
  \begin{tabular}{lcc}
    \toprule
     & WER(\%) $\downarrow$ \\
    \midrule
    \url{(A1)}CTC (greedy) & $18.10$ \\
    \url{(A2)}+EC (MLM, $\alpha = 1$) & $18.48$ \\
    \url{(A3)}+EC (MLM, $\alpha = 0.3$) & $18.02$ \\
    \url{(A4)}+EC (PC-MLM, $\alpha = 1$) & $17.54$ \\
    \url{(A5)}+EC (PC-MLM, $\alpha = 0.5$) & $16.94$ \\
    \url{(A6)}+EC (Del PC-MLM, $\alpha = 1$) & $16.82$ \\
    \url{(A7)}+EC (Del PC-MLM, $\alpha = 0.5$) & $\bm{16.48}$ \\
    \url{(A8)}+phone-to-word CTC & $23.85$ \\ \hline
    \url{(B1)}CTC Conformer (greedy) & $16.53$ \\
    \url{(B2)}+EC (Del PC-MLM, $\alpha = 0.5$) & $\bm{15.48}$ \\
    \bottomrule
 \end{tabular}
\end{table}
\endgroup


Table \ref{tab:ec-csj-resc-sf} compares our method on CSJ with other LM integration methods.
Real time factor (RTF) are noted in the table, which was measured with a batch size of $1$ using an NVIDIA TITAN V GPU by averaging five runs.
PC-MLMs increased RTF compared to MLM because it utilizes phone tokens as input which are longer than word tokens in general.
Our method worked much faster, compared to rescoring (Resc) \url{(C2, C3)} and shallow fusion (SF) \url{(C4)}, which require beam search \url{(C1)}. In shallow fusion, LM calculation at each decoding step is also required.
Note that beam search is hard to parallelize, but Transformer inference in our error correction benefits from parallelization with GPU.

We also compared our method to knowledge distillation (KD) with MLM.
To apply KD, we utilized a forced-aligned CTC path to align frame-level predictions of CTC with token-level predictions of MLM, inspired by \cite{Inaguma21-AKD}.
The KD loss function is formulated as
\begin{align}
\mathcal{L}_{\rm KD} = - \frac{1}{\sum_{i=1}^L |\mathcal{A}(i)|} \sum_{i=1}^{L} \sum_{t \in \mathcal{A}(i)} \sum_{v \in \mathcal{V}} P_{\rm MLM}^{(i, v)} \log P_{\rm CTC}^{(t, v)},
\end{align}
where $\mathcal{A}$ denotes the mapping from $i$ to $t$ based on the forced-aligned CTC path calculated with CTC forward-backward algorithm \cite{Graves06-CTC}.
KD was applied during training, so its RTF was not increased from the baseline.
However, its WER improvement was limited \url{(D1)} compared to our method.
The combination of KD and our method \url{(D2)} leads to further WER improvement.
The WER was improved by $10.7\%$ relative over the baseline, while maintaining fast inference.

\begingroup
\renewcommand{\arraystretch}{1.1}
\begin{table}[t]
  \caption{Comparison with other LM integration methods on CSJ-APS: shallow fusion (SF), rescoring (Resc), knowledge distillation (KD). ``BS'' means beam search without LM, and $b$ denotes beam width. $n$ denotes the number of hypotheses to rescore.}
  \label{tab:ec-csj-resc-sf}
  \centering
  \begin{tabular}{lcc}
    \toprule
     & WER(\%) $\downarrow$ & RTF $\downarrow$ \\
    \midrule
    \url{(A1)}CTC {(greedy)} & $18.10$ & $0.0033$ \\
    \url{(A3)}+EC {(MLM)} & $18.02$ & $0.0067$ \\
    \url{(A7)}+EC {(Del PC-MLM)} & $16.48$ & $0.0094$ \\ \hline
    \url{(C1)}CTC+BS {($b=5$)} & $18.08$ & $0.043$ \\
    \url{(C2)}+Resc {(TLM, $n=5$)} & $17.22$ & $0.052$ \\
    \url{(C3)}+Resc {(MLM, $n=5$)} & $17.20$ & $0.22$ \\
    \url{(C4)}+SF {(TLM, $b=5$)} & $16.51$ & $0.38$ \\ \hline
    \url{(D1)}CTC+KD {(MLM)} & $17.44$ & $0.0033$ \\
    \url{(D2)}+EC {(Del PC-MLM)} & $\bm{16.16}$ & $0.0094$ \\
    \bottomrule
 \end{tabular}
\end{table}
\endgroup

Table \ref{tab:ec-ted2} shows the ASR results of our proposed error correction (EC) on TED-LIUM2.
Our method improved ASR, by taking advantage of LM, while maintaining fast inference, as on CSJ.
However, in terms of WER, our method was not competitive with rescoring (Resc) \url{(F2)} and shallow fusion (SF) \url{(F3)}.
We found that phone information was not as effective as on CSJ, by comparing \url{(E2)} and \url{(E3)}.
English words (alphabet) are a {\it phonogram} that represents speech sound while Japanese words (kanji) are an {\it ideogram} that represents a concept (meaning).
Therefore, phone information is more related to word information in English than in Japanese.
This suggests that word-level recognition errors and phone-level ones are likely to occur at the same position and that phone information was not helpful in error correction.
On the other hand, phone information had a complementary effect on Japanese words.

\begingroup
\renewcommand{\arraystretch}{1.1}
\begin{table}[t]
  \caption{ASR results on TED-LIUM2 with proposed error correction (EC). ASR models were trained on Librispeech, and LMs were trained on TED-LIUM2 text.}
  \label{tab:ec-ted2}
  \centering
  \begin{tabular}{lcc}
    \toprule
     & WER(\%) $\downarrow$ & RTF $\downarrow$ \\
    \midrule
    \url{(E1)}CTC (greedy) & $17.99$ & $0.0021$ \\
    \url{(E2)}+EC (MLM) & $17.62$ & $0.0045$ \\
    \url{(E3)}+EC (PC-MLM) & $17.59$ & $0.0059$ \\ 
    \url{(E4)}+EC (Del PC-MLM) & $\bm{17.35}$ & $0.0059$ \\ \hline
    \url{(F1)}CTC+BS ($b=5$) & $17.98$ & $0.035$ \\
    \url{(F2)}+Resc (TLM, $n=5$) & $16.67$ & $0.041$ \\
    \url{(F3)}+SF (TLM, $b=5$) & $\bm{15.91}$ & $0.37$ \\ \hline
    \url{(G1)}CTC+KD (MLM) & $17.66$ & $0.0021$ \\
    \url{(G2)}+EC (Del PC-MLM) & $\bm{17.02}$ & $0.0059$ \\
    \bottomrule
 \end{tabular}
\end{table}
\endgroup

\vspace{-2pt}
\section{Conclusions}
\vspace{-2pt}
In this study, we have proposed an LM integration method for CTC-based ASR via error correction with phone-conditioned masked LM (PC-MLM).
PC-MLM corrects less confident tokens in CTC output using phone information.
We demonstrated our proposed method worked faster than conventional LM integration methods such as rescoring and shallow fusion.
They assume multiple hypotheses from autoregressive beam search, while our method assumes a single hypothesis from non-autoregressive greedy decoding.
In addition, PC-MLM itself works in a non-autoregressive manner.
We also demonstrated on CSJ that our method even improved the ASR performance more than rescoring, shallow fusion, and knowledge distillation.
For future work, we will investigate recovering deletion errors and iterative refinement in PC-MLM, while keeping its inference speed.

\bibliographystyle{IEEEtran}
\bibliography{mybib}

\end{document}